%% file: aaai25.tex
\documentclass[letterpaper]{article} 
\usepackage{aaai25}  
\usepackage{times}  
\usepackage{helvet}  
\usepackage{courier}  
\usepackage[hyphens]{url}  
\usepackage{graphicx} 
\def\UrlFont{\rm}  
\usepackage{natbib}  
\usepackage{caption} 
\usepackage{algorithm}
\usepackage{algorithmic}
\usepackage{amsmath}
\usepackage{booktabs}
\usepackage{makecell}

\usepackage{newfloat}
\usepackage{listings}
\DeclareCaptionStyle{ruled}{labelfont=normalfont,labelsep=colon,strut=off} 
\floatstyle{ruled}
\newfloat{listing}{tb}{lst}{}
\floatname{listing}{Listing}
\title{SparseTem: Boosting the Efficiency of CNN-Based Video Encoders by Exploiting Temporal Continuity}
\author{
    Kunyun Wang\textsuperscript{\rm 1},
    Shuo Yang\textsuperscript{\rm 2},
    Jieru Zhao\thanks{Corresponding author: Jieru Zhao (zhao-jieru@sjtu.edu.cn)}\textsuperscript{\rm 1},
    Wenchao Ding\textsuperscript{\rm 3},
    Quan Chen\textsuperscript{\rm 1},
    Jingwen Leng\textsuperscript{\rm 1},
    Minyi Guo\textsuperscript{\rm 1}
}
\affiliations{
    \textsuperscript{\rm 1}School of Computer Science, Shanghai Jiao Tong University \\
    \textsuperscript{\rm 2}Department of Electrical Engineering and Computer Sciences, University of California, Berkeley \\
    \textsuperscript{\rm 3}College of Intelligent Robotics and Advanced Manufacturing, Fudan University \\
    wkykaixin@sjtu.edu.cn, andy\_yang@berkeley.edu, zhao-jieru@sjtu.edu.cn, dingwenchao@fudan.edu.cn, chen-quan@cs.sjtu.edu.cn, leng-jw@sjtu.edu.cn, guo-my@cs.sjtu.edu.cn
}

\usepackage{bibentry}

\nocopyright

\begin{document}

\maketitle

\begin{abstract}
Deep learning models have become pivotal in the field of video processing and is increasingly critical in practical applications such as autonomous driving and object detection. Although Vision Transformers (ViTs) have demonstrated their power, Convolutional Neural Networks (CNNs) remain a highly efficient and high-performance choice for feature extraction and encoding. However, the intensive computational demands of convolution operations hinder its broader adoption as a video encoder. Given the inherent temporal continuity in video frames, changes between consecutive frames are minimal, allowing for the skipping of redundant computations. This technique, which we term as Diff Computation, presents two primary challenges. First, Diff Computation requires to cache intermediate feature maps to ensure the correctness of non-linear computations, leading to significant memory consumption. Second, the imbalance of sparsity among layers, introduced by Diff Computation, incurs accuracy degradation. To address these issues, we propose a memory-efficient scheduling method to eliminate memory overhead and an online adjustment mechanism to minimize accuracy degradation. We integrate these techniques into our framework, SparseTem, to seamlessly support various CNN-based video encoders. SparseTem achieves speedup of 1.79x for EfficientDet and 4.72x for CRNN, with minimal accuracy drop and no additional memory overhead. Extensive experimental results demonstrate that SparseTem sets a new state-of-the-art by effectively utilizing temporal continuity to accelerate CNN-based video encoders. 
  
\end{abstract}

%

\input{sec-intro}
\input{sec-related}

\input{sec-method}
\input{sec-experiments}
\input{sec-conclusion}

\clearpage %



\bibliography{aaai25}
\end{document}

%% file: sec-intro.tex
\section{Introduction}
Deep learning models have achieved remarkable success in video processing, including video recognition \cite{twostream,deep-feature-flow}, object detection \cite{faster-r-cnn,hrnet,Efficientdet,yolo,ssd} and autonomous driving \cite{surroundocc,occformer,openoccupancy}. 
In these tasks, models generally consist of an encoder and a decoder. The encoder extracts features, and the decoder produces the final output. Encoders, often implemented as deep Convolutional Neural Networks (CNNs), become bottlenecks due to their computational intensity, requiring billions of FLOPS. This has led to a strong demand for accelerating CNN-based video encoders.
\begin{figure} 
  \centering
  \includegraphics[width=\linewidth]{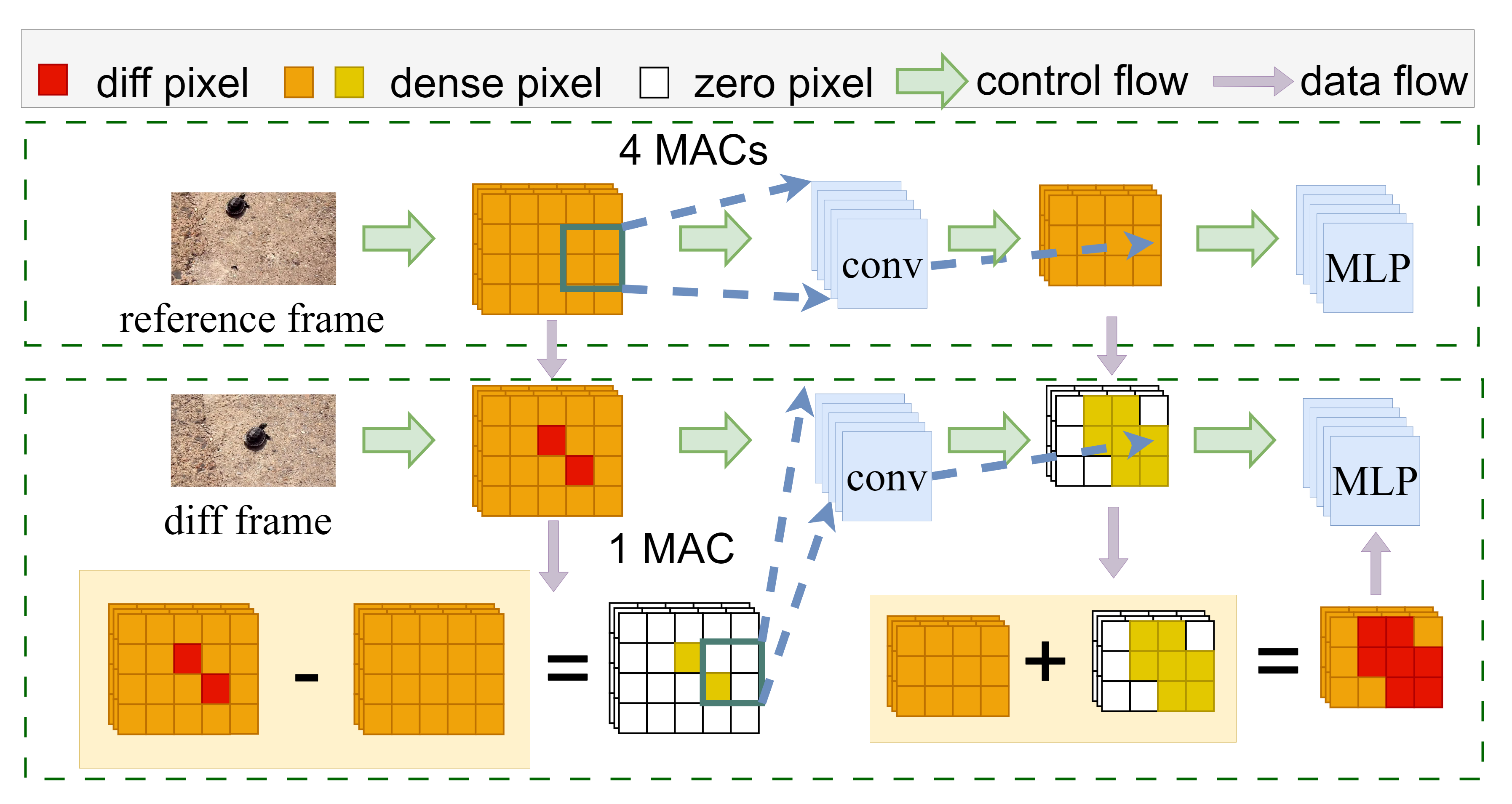}
  \vspace{-20pt}
  \caption{Illustration for Diff Computation on a toy model consisted with only one convolutional layer and one MLP layer. Dense computation is performed on reference frame, which is usually the first frame of a video chunk,
  as shown in the first row. Diff Computation is performed on subsequent frames, i.e., diff frames, which only computes on diff pixels and skip the computation for zero pixels, as shown in the second row.
  Frame taken from \cite{imagenet}. 
  }
  \label{fig:diff_computation}
\end{figure}
From a general perspective, researchers have explored multiple avenues to reduce the computational burden and accelerate CNNs. Some efforts focus on architectural design, creating custom hardware \cite{Eyeriss,SCNN,Cambricon-X}; others concentrate on software, developing custom libraries and operators \cite{tvm,unified,minuet}; and some approaches employ algorithmic strategies such as pruning \cite{hanlearning,lipruning,anwar2017structured} and quantization \cite{hubara2018quantized,jacob2018quantization} to decrease computational costs and achieve acceleration.

Though these methods achieve efficiency improvement for general CNNs, they have not considered the intrinsic property of videos, i.e., temporal continuity, which can further increase the speedup. 
To leverage the temporal redundancy in video encoders, prior arts propose various efficient CNN architectures. Several works only involve backbone computations on key-frames \cite{twostream, deep-feature-flow, accel} or use shallow backbones in an interleaved manner \cite{fast-slow, dynamic-kernel}. And some methods reduce the computational burden of 3D convolutions which are popular in video processing \cite{closer, tsm}. Specially designed CNN architectures are efficient, however, the techniques are not general to other models.

\begin{figure}[t] 
  \centering
  \includegraphics[width=\linewidth]{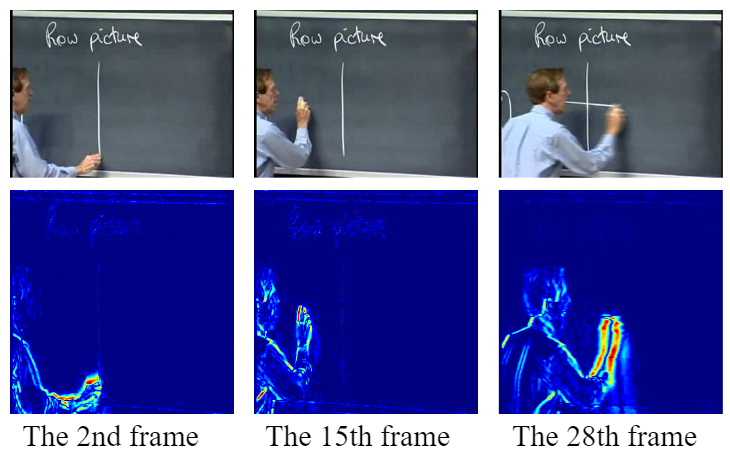}
  \vspace{-20pt}
  \caption{Illustration of the magnitude of pixel differences
 between consecutive frames of a video chunk. Frames taken from  \cite{ucf101}.}
  \label{fig:heatmap}
\end{figure}

There exist more general methods leveraging the temporal continuity of videos from the perspective of computation to minimize redundancy in CNNs \cite{skip-conv,deltacnn,cbinfer,tempdiff}. 
They utilize the temporal continuity of videos to perform convolutions and other calculations only on pixels or regions that have changed, thereby skipping redundant computations. We term this techniques as Diff Computation, as illustrated in Figure \ref{fig:diff_computation}. While theoretically reducing CNN computational load, these methods \cite{skip-conv,cbinfer,tempdiff} introduce additional control overhead and lead to unstructured data, which modern GPUs, designed with SIMD architecture, cannot handle efficiently. Unstructured data also adversely affects memory access efficiency. Therefore, translating theoretically reduced computational load into practical acceleration is challenging.

DeltaCNN \cite{deltacnn} designs sparse operators, using a mask to skip redundant computations at the pixel level, achieving efficiency improvement compared to cuDNN. However, to address non-linear issues, DeltaCNN requires caching intermediate feature maps proportional to non-linear layers, resulting in significant memory overhead which makes it not suitable for deep CNNs.

\textit{None of the previous methods simultaneously address acceleration, memory overhead, and accuracy. While \cite{skip-conv} theoretically reduces the amount of computation, it does not result in actual acceleration. \cite{deltacnn} achieves real acceleration, but its significant memory overhead hinders its application. \cite{tempdiff} uses diff computation in only some layers, which reduces additional memory overhead but also diminishes the acceleration effect.}

In this paper, we introduce SparseTem, a computationally efficient and memory-friendly framework for Diff Computation. SparseTem effectively addresses the memory overhead challenges posed by previous approaches and incorporates a novel online adjustment mechanism to minimize accuracy degradation in Diff Computation. SparseTem can simultaneously address acceleration, memory overhead, and accuracy, making it generalizable to any CNN-based video encoder.
The main contributions of this paper are:

\begin{itemize}
    \item We propose a computationally efficient and memory-friendly framework for Diff Computation, SparseTem, which can efficiently accelerate CNN-based video encoders with no additional memory overhead and minimal accuracy degradation. Our framework is compatible with various CNNs with slight adjustment.
    \item We introduce a novel scheduling method, SparseBatch, which eliminates dependency conflicts between adjacent frames in a video chunk and utilizes temporal continuity for Diff Computation. By passing intermediate feature maps through the chunk, our method removes additional caching, thereby decreasing memory overhead.
    \item Our study reveals substantial variations in the sparsity levels of intermediate feature maps across different layers when utilizing Diff Computation. This imbalance in sparsity results in significant accuracy degradation. To mitigate this issue, we introduce a novel online adjustment mechanism that effectively eliminates the imbalance and minimizes accuracy degradation.
\end{itemize}

Experimental results show that SparseTem achieves up to 4.72x end-to-end speedup with minimal accuracy loss compared to cuDNN. Additionally, SparseTem significantly reduces memory overhead, lowering it by 68.1\% compared to DeltaCNN and by 28.6\% compared to cuDNN.


%% file: sec-related.tex
\section{Related Work}

\begin{figure*}[t]
  \centering
  \includegraphics[width=\textwidth]{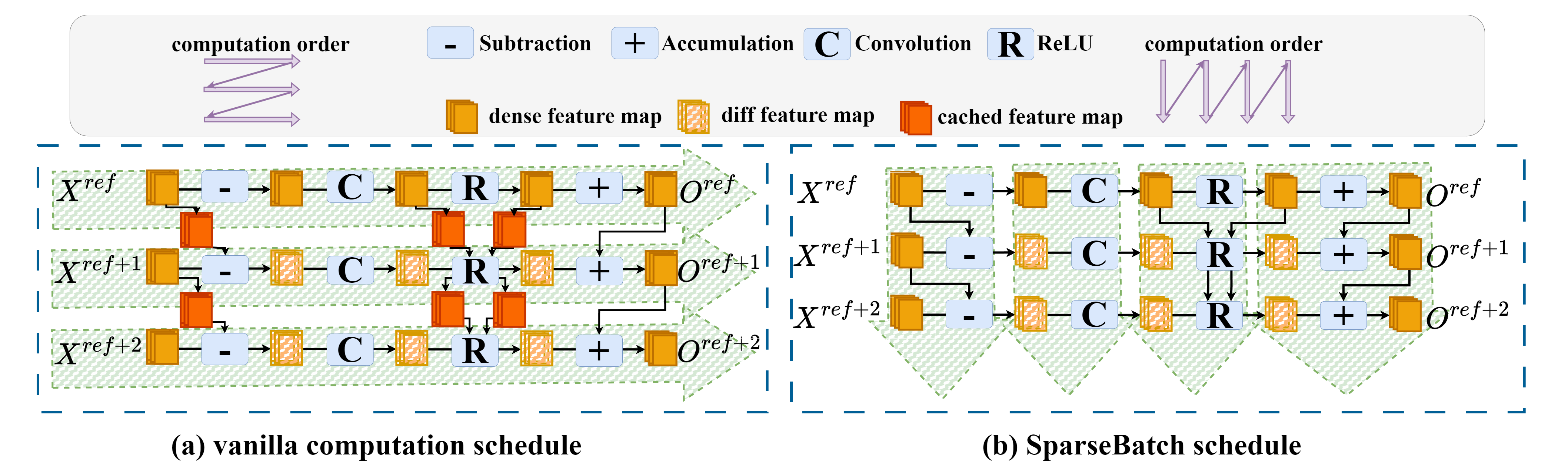}
  \vspace{-20pt}
  \caption{Illustration of vanilla computation schedule and SparseBatch schedule. $X^{ref}$ is the first frame in a video chunk. SparseTem performs dense computation on $X^{ref}$ while performs Diff Computation on $X^{ref+1}$ and $X^{ref+2}$. 
  The purple arrows denote computation order.
  }
  \label{fig:normal_sparsebatch_computation}
\end{figure*}
\subsection{Efficient CNN based video encoders}
Researchers have developed efficient CNN architectures for video encoding. Some approaches reduce computational costs by applying expensive backbone computations only to key frames, performing spatial computations on key frames and temporal computations on multi-frame optical flow \cite{twostream,deep-feature-flow,accel}. However, the density of optical flow can make temporal computation a bottleneck. To further reduce costs, shallow backbones have been proposed to work in an interleaved manner, extracting features from non-key frames, while deep backbones are used only for key frames \cite{fast-slow,dynamic-kernel}. These features are fused via concatenation or RNNs to produce the final output. 3D CNNs are widely popular in video processing, and efforts have been made to reduce the computational burden of 3D convolutions. For instance, Temporal Shift Module \cite{tsm}, which shifts feature maps along the temporal dimension, preserving temporal information while reducing computation. Additionally, other works \cite{closer} analyze 3D convolutions, proposing optimizations that lower computation costs while maintaining high accuracy in action recognition tasks.

Although specially designed model architectures boost efficiency of video encoders, it is challenging to transfer this efficiency to other models. SparseTem leverages temporal continuity from the perspective of computation. It skips redundant computations in CNNs while maintaining high accuracy and low memory overhead, and can be applied to any CNN-based video encoder.

\subsection{Utilizing Temporal Redundancy in Videos}
Temporal continuity is an intrinsic property of videos, particularly in those captured by surveillance cameras and license plate recognition cameras. Several studies, including Skip-conv \cite{skip-conv}, Cbinfer \cite{cbinfer}, Tempdiff \cite{tempdiff}, and DeltaCNN \cite{deltacnn}, have leveraged this property to reduce computational redundancy by performing operations only on changed pixels or regions. Skip-conv uses a gating network to bypass redundant calculations but lacks the implementation of sparse computation operators, limiting its practical acceleration. Cbinfer accelerates convolutions by caching previous frame data but incurs significant memory and computational overhead, making it unsuitable for large networks. Tempdiff conserves memory by limiting redundant computation skipping to the first few layers, thereby limiting performance gains. DeltaCNN introduces methods to handle non-linear computations like ReLU by restoring sparse features to their original form, but this approach increases memory overhead, making it less efficient for commercial-grade GPUs with limited memory capacity.

%% file: sec-method.tex
\section{Method}
Figure \ref{fig:heatmap} shows the heatmaps of pixel differences between consecutive frames, indicating the sparsity of values. For the UCF101 \cite{ucf101} dataset, the differences between adjacent frames exhibit a sparsity greater than 50\%, allowing Diff Computation to achieve at least a 50\% reduction in computations. Factors such as camera jitter and noise contribute to many small differences between frames. By applying a threshold to filter out non-zero pixels below this threshold, the sparsity can be further enhanced.


\subsection{Skipping Non-zero Delta Pixels}
The majority of video encoders either contain CNN components or are fully CNN-based models. Conventional CNNs conduct dense computations on all the pixels of each input frame. In contrast, SparseTem performs dense computations exclusively on the \textit{reference frame}, usually the first frame of a video chunk. For subsequent frames, known as \textit{diff frames}, SparseTem uses Diff Computation. 

In Diff Computation (Figure \ref{fig:diff_computation}), the reference frame is subtracted from the current frame to obtain sparse feature maps, a process referred to as 
\textit{Subtraction}. These sparse feature maps primarily consist of zero pixels, where all channels are zero, and a few diff pixels, where at least one channel is non-zero. Convolutions are performed solely on these diff pixels. Subsequently, the generated sparse feature maps are integrated with the output dense feature maps of the reference frame, producing the output dense feature maps for the current frame, a process called \textit{Accumulation}. The linear properties of convolution operations ensure the correctness of Diff Computation.

\subsubsection{Linear property of convolution}
Consider a convolutional layer with a kernel $W \in \mathbf{R}^{c_o \times c_i \times k_h \times k_w}$, an input feature map $X \in \mathbf{R}^{c_i \times h_i \times w_i}$, and an output feature map $O \in \mathbf{R}^{c_o \times h_o \times w_o}$. The original convolution is defined as:
\begin{equation}
O_{i,j,k} = \sum_{m=1}^{c_i} \sum_{n=1}^{k_h} \sum_{p=1}^{k_w} W_{i,m,n,p} \cdot X_{m, j+n-1, k+p-1},
\end{equation}
where $i = 1, \ldots, c_o$, $j = 1, \ldots, h_o$, and $k = 1, \ldots, w_o$. If we consider the Diff Computation: 
\begin{equation}
\begin{aligned}
    O &= W \times X \\
      &= W \times (X^{ref} + X - X^{ref}) \\
      &= W \times X^{ref} + W \times X^{\Delta} \\
      &= O^{ref} + W \times X^{\Delta},
\end{aligned}
\end{equation}
where $X^{ref}$ and $O^{ref}$ denote input and output feature maps of the reference frame, respectively. Diff Computation only requires processing $X^{\Delta}$ to obtain the output $O$. Since pixels in $X^{\Delta}$ are predominantly zero, the majority of computation can be skipped, thus greatly improving inference speed. However, it is important to note that not all computations are linear. Non-linear computations, such as most activation functions in CNNs, also need to be considered.

\subsubsection{Non-linear operations} Most activation functions are non-linear operations. Applying non-linear operations to sparse feature maps can result in erroneous outcomes, i.e.
\begin{equation}
O \neq O^{ref} + f_{non-linear}(X^{\Delta})
\end{equation}
To ensure the correctness of Diff Computation, DeltaCNN \cite{deltacnn} proposes a solution based on caching intermediate feature maps, as illustrated in Figure \ref{fig:normal_sparsebatch_computation}(a). DeltaCNN utilizes a buffer to cache the input and output dense feature maps of non-linear operations for each layer. During Diff Computation, the cached input feature map is used to reconstruct the dense feature map from the sparse input feature map before the non-linear operation. Non-linear operation is then operated on the reconstructed dense feature map. After the non-linear operation, the cached output feature map is used to restore the sparse feature map from the dense output feature map produced by the non-linear operation. The buffer updates the cached feature maps during this process. Although DeltaCNN effectively addresses the errors caused by non-linear operations in Diff Computation, caching intermediate feature maps results in significant memory overhead, which is proportional to the number of non-linear layers. This memory overhead limits its application in memory-constrained devices and edge scenarios. We propose SparseBatch to eliminate this memory overhead. 
%

\subsubsection{Truncation threshold}
Due to noise in the frames, there are often many minor changes between adjacent frames in the video. Meanwhile, convolution can cause \textit{dense amplification phenomenon}. For instance, if there is only one non-zero pixel in a sparse feature map, a 3x3 convolution will produce nine non-zero pixels in the sparse feature map. To avoid redundant computation on these minor changes and mitigate the impact of dense amplification phenomenon, we need to truncate small values in the feature maps. DeltaCNN \cite{deltacnn} truncates pixels using a fixed truncation threshold. If the value of all channels of a pixel is smaller than this threshold, it will be truncated to zero. Conversely, SparseTem utilizes a novel online adjustment mechanism to dynamically adjust truncation thresholds.

\subsection{Design of SparseBatch}
\label{sec:SparseBatch}
To eliminate the memory overhead caused by caching intermediate feature maps for non-linear operations, we propose an innovative scheduling method called SparseBatch. This method resolves dependency conflicts between adjacent frames within a video chunk. By passing intermediate feature maps, SparseBatch eliminates the need for additional caching, as illustrated in Figure \ref{fig:normal_sparsebatch_computation}(b). 

For Diff Computation, placing adjacent frames from a video chunk into a batch is a non-trivial task. In the vanilla computation schedule, Subtraction produces sparse feature maps based on the previous batch. Since Diff Computation relies on the similarity between adjacent frames, placing them within a single batch can reduce the similarity between adjacent batches, potentially increasing the number of non-zero pixels in the sparse feature maps. To minimize redundant computations while batching adjacent frames, it is necessary to resolve dependency conflicts between them.

SparseBatch performs dense computations on the first frame of each video chunk (noting that a batch may contain multiple video chunks) and initializes the buffers for Subtraction and Accumulation. Since the model keeps only one buffer for Subtraction and one for Accumulation, the memory overhead remains negligible. Subsequent frames in the batch are processed as diff frames. SparseBatch uses Subtraction to sequentially generate sparse feature maps for each diff frame. After Subtraction, dense computations are applied to the dense feature maps, while Diff Computation is applied to the sparse feature maps for each linear operation. For non-linear operations, SparseBatch processes each frame individually, propagating the input and output feature maps to the next frame. This approach effectively eliminates memory overhead. In summary, SparseBatch changes the computation order from a "Z" shape to an "N" shape, optimizing the process to handle dependency conflicts and reduce memory usage.

\begin{figure} 
  \centering
  \includegraphics[width=\linewidth]{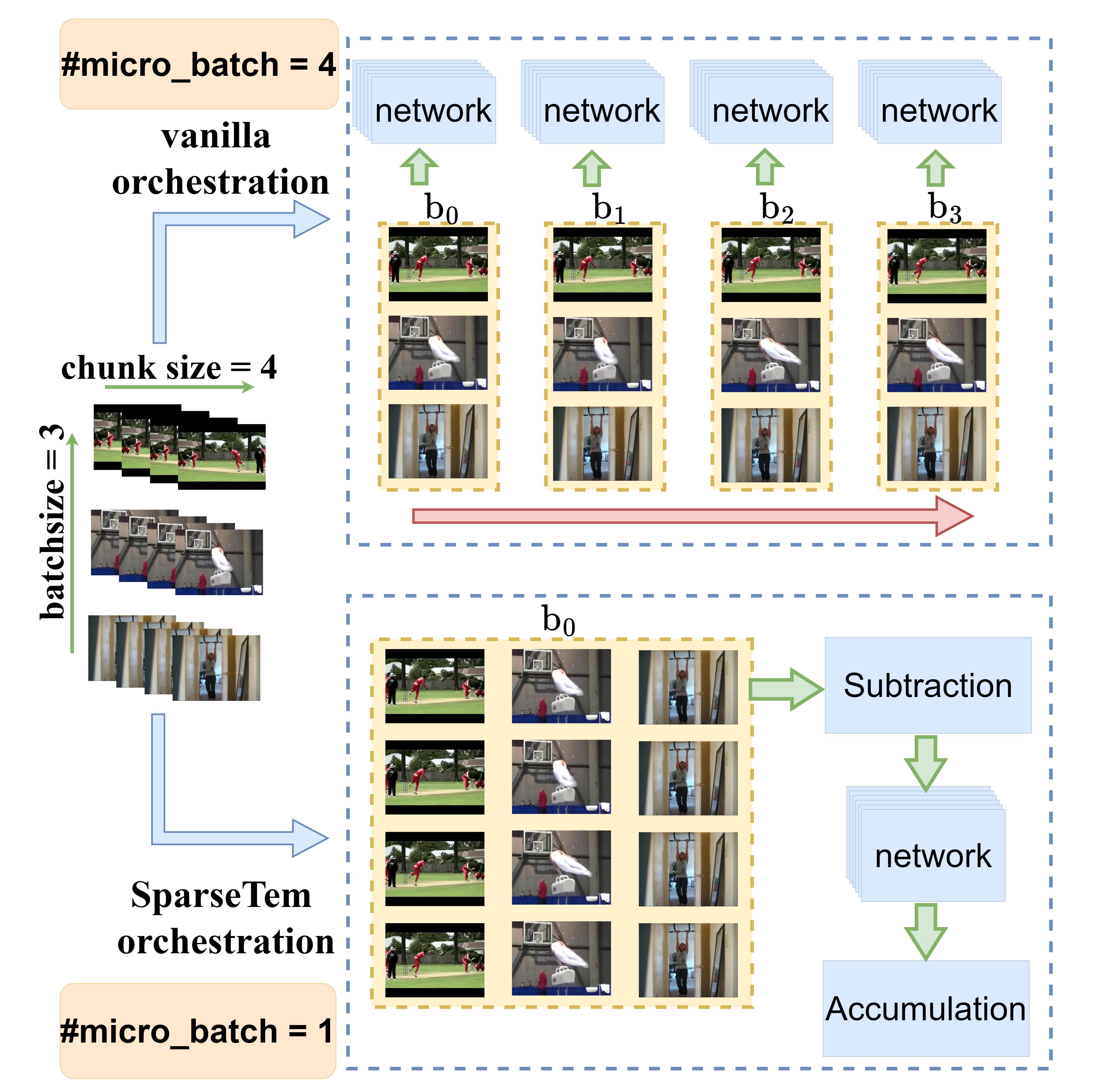}
  \vspace{-20pt}
  \caption{This figure shows the inference process over three videos with a chunk size of four. The vanilla orchestration method requires inference on four micro-batches, whereas SparseTem needs inference on only a single micro-batch.}
  \label{fig:orchestration}
\end{figure}

\subsection{Data Orchestration}
\label{sec:orchestration }
Video encoders typically process multiple videos in a single inference pass, with batch size determined by the number of videos and the chunk size of video segments. Traditional Diff Computation cannot handle adjacent video frames within a single batch simultaneously. 
Methods such as DeltaCNN \cite{deltacnn} divide the batch into multiple micro batches, each handling the i-th frame and requiring as many inferences as the chunk size. In contrast, SparseBatch removes the dependency between adjacent frames, enabling them to be processed in a single batch. As a result, SparseTem requires only one inference pass. Figure \ref{fig:orchestration} compares traditional and SparseTem orchestration.

\subsection{Online Truncation Threshold Adjustment}
\label{sec:Online}

We observed that using a fixed truncation threshold in SparseTem results in significant sparsity imbalances across different layers, as shown in Figure \ref{fig:sparsity_diff_fixed_online}(a). This imbalance prevents optimal acceleration in layers with lower sparsity and causes accuracy degradation in layers with higher sparsity. To address this, we propose an Online Truncation Threshold Adjustment mechanism, which dynamically adjusts the threshold for each layer to align the sparsity of intermediate feature maps with a target range. By searching for thresholds around a target value $T$ within a range $[T-\epsilon, T+\epsilon]$, this mechanism helps balance sparsity and maintain performance across all layers.

To be more specific, we propose an iterative binary search for threshold optimization.
The Online Truncation Threshold Adjustment mechanism dynamically adjusts the truncation threshold based on the sparsity of intermediate feature maps in nonlinear layers (e.g., activation layers). To minimize the overhead of online adjustment, we initially proposed Binary Search for Threshold Optimization (BST) which can find suitable threshold quickly. BST adjusts the threshold by evaluating the sparsity from each sample and determining whether to increase or decrease the threshold. However, BST performed poorly due to significant variations in sparsity across different inputs (see Figure \ref{fig:sparsity_diff_fixed_online}(b)), making it unsuitable for all cases.

To address this, we developed Iterative Binary Search for Threshold Optimization (IBST). IBST performs a BST in each cycle, allowing it to periodically adjust the threshold to accommodate varying inputs. This iterative approach effectively balances sparsity and reduces accuracy degradation, as demonstrated by our experiments. 


\subsection{Implementation}
SparseTem is a memory-efficient and computation-efficient framework built upon DeltaCNN \cite{deltacnn}. DeltaCNN provides CUDA kernels and PyTorch extensions for Diff Computation. SparseTem extends these kernels and PyTorch extensions to support the SparseBatch scheduling strategy and Online Truncation Threshold Adjustment.

\begin{figure}[ht] 
  \centering
  \includegraphics[width=\linewidth]{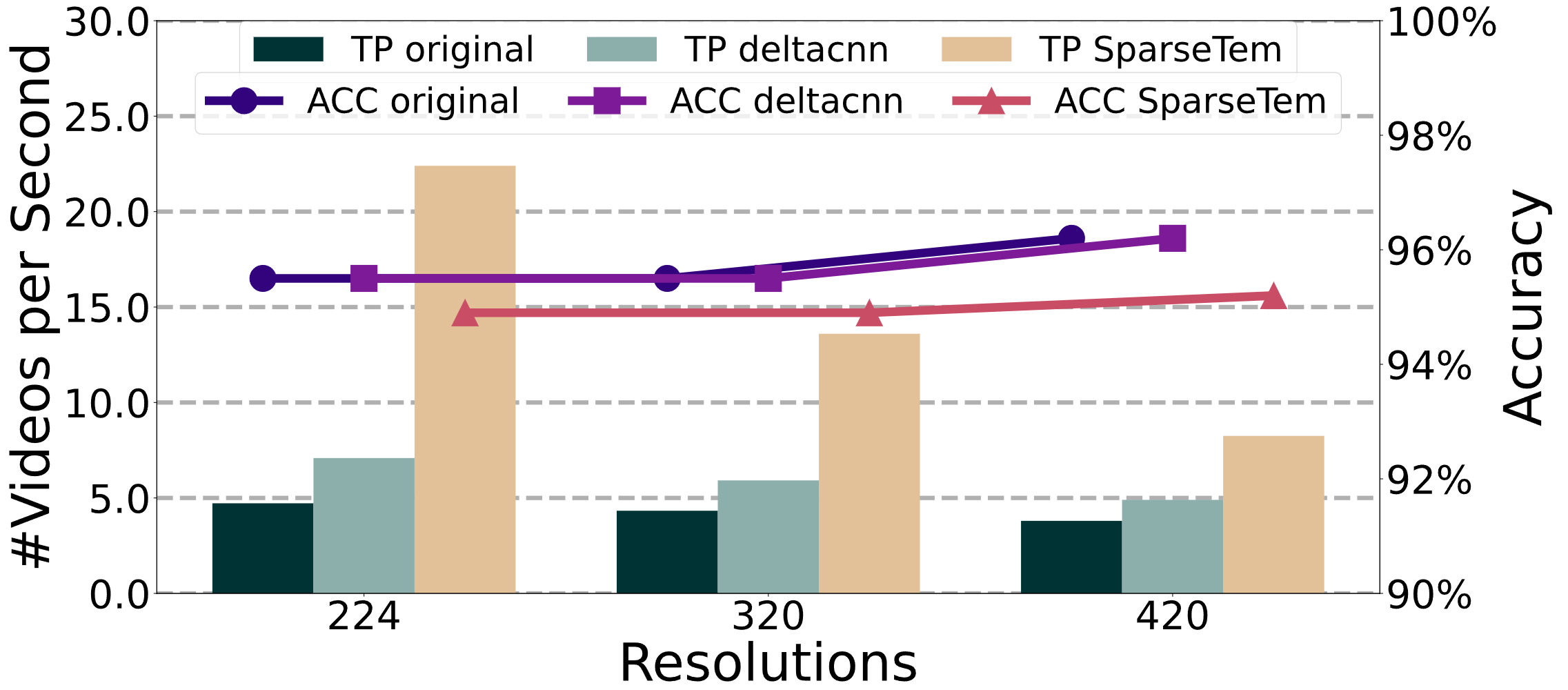}
  \vspace{-20pt}
  \caption{Throughput (TP) and accuracy (ACC) of SparseTem compared with original model and DeltaCNN.}
  \label{fig:crnn_end2end}
\end{figure}

%% file: sec-experiments.tex
\section{Experiments}

\begin{figure*}[ht] 
  \centering
  \includegraphics[width=\linewidth]{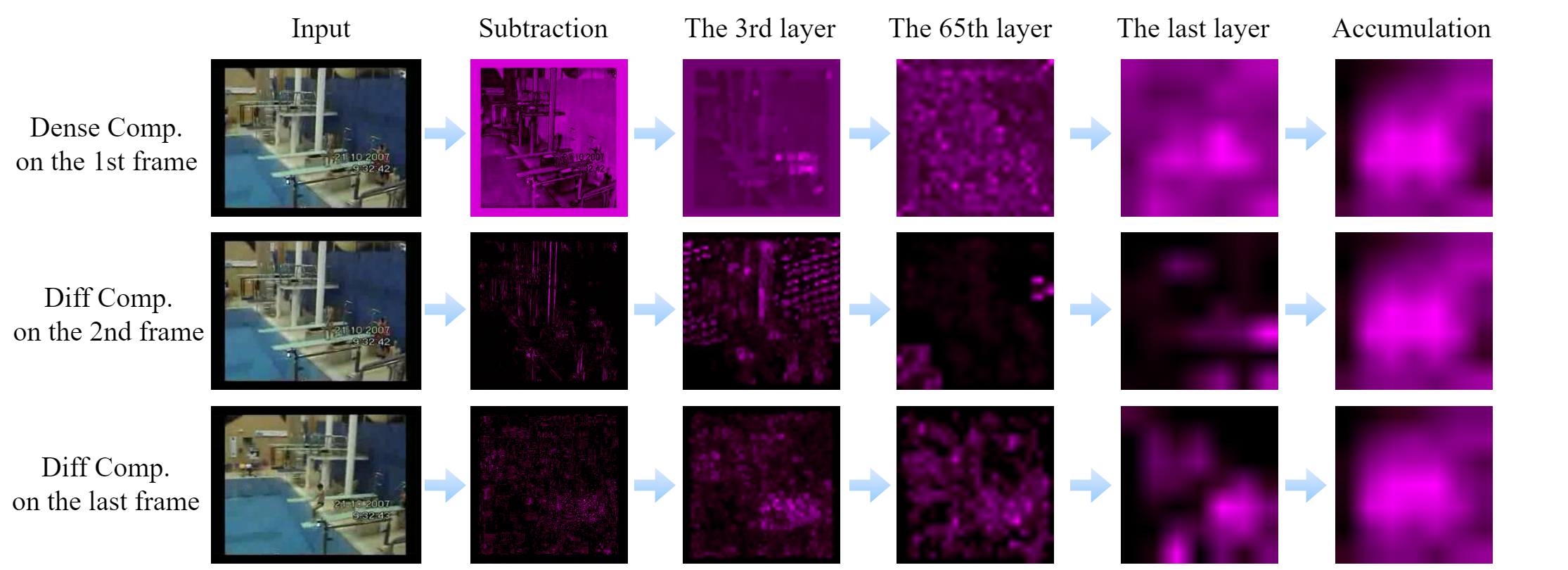}
  \caption{
    Feature map visualization of CRNN+SparseTem during inference on a video chunk. SparseTem skips a large amount of computation on the pixels that are black and still produces the correct output after Accumulation. 
  }
  \label{fig:crnn_demo}
\end{figure*}

\begin{figure*}[ht] 
  \centering
  \includegraphics[width=\linewidth]{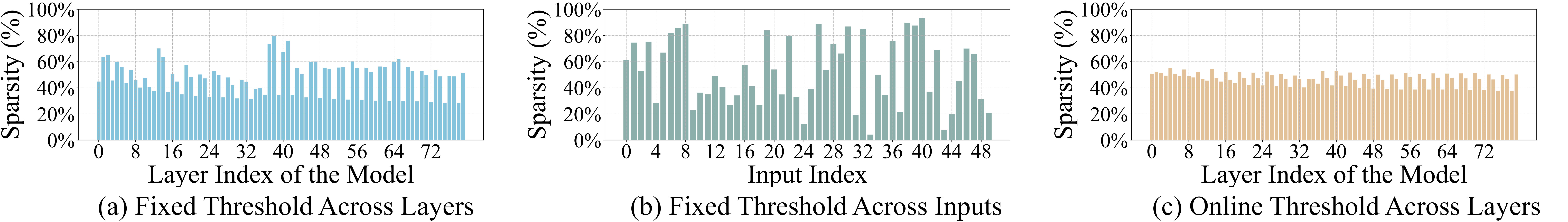}
  \caption{
  (a) Sparsity varies significantly across layers for the same input. (b) Sparsity varies significantly within the same layer for different inputs. (c) Online Threshold Truncation Adjustment balances sparsity across layers.
  }
  \label{fig:sparsity_diff_fixed_online}
\end{figure*}

In this section, we demonstrate the superiority of the proposed SparseTem framework through extensive experiments on UCF101 \cite{ucf101} action recognition dataset and MOT16 \cite{mot16} object detection dataset. For UCF101, we used CRNN \cite{crnn} whose backbone is ResNet152 \cite{resnet}. For MOT16, we used Efficientdet \cite{Efficientdet} whose backbone is Efficientnet \cite{efficientnet}. We trained both the networks on video datasets using pre-trained model weights from image datasets. 
The experiments in this study were conducted on a server equipped with a 48-core Intel(R) Xeon(R) Silver 4310 CPU (2.10GHz) and two NVIDIA 4090 GPUs (24GB VRAM each). The CUDA version used was 11.3, and the PyTorch version was 1.10.1.

\subsection{Action Recognition}

\begin{figure}[ht] 
  \centering
  \includegraphics[width=\linewidth]{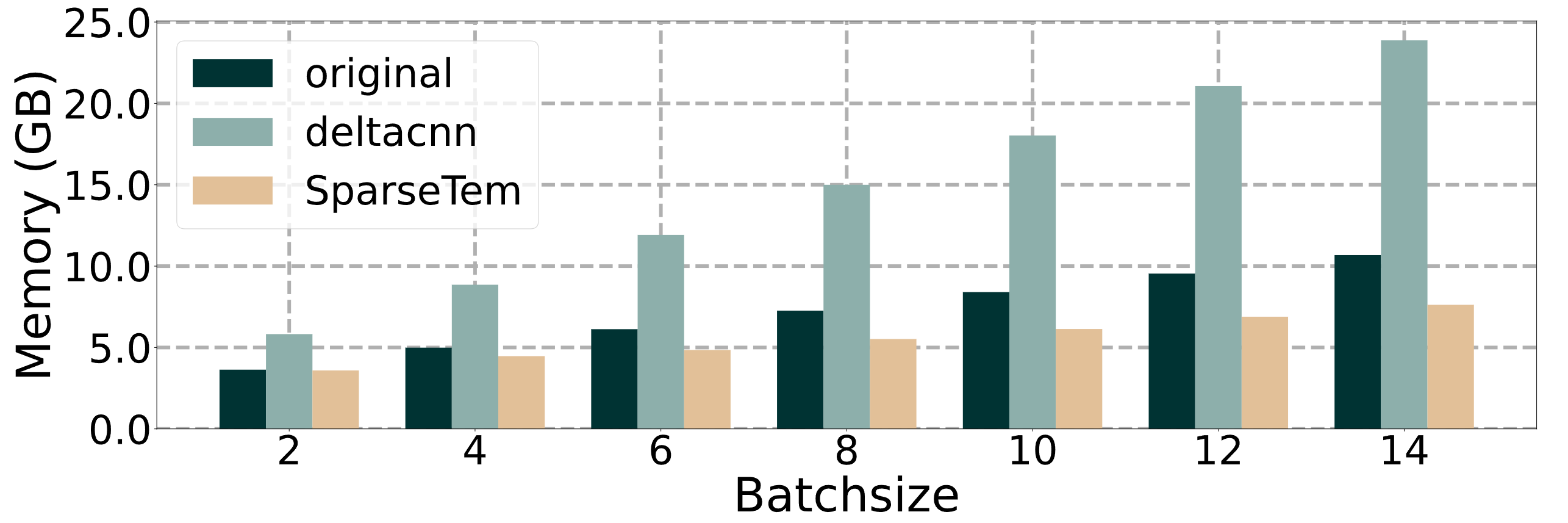}
  \caption{Comparison memory cost of  SparseTem, original model, and DeltaCNN on Efficientdet-d3.}
  \label{fig:memory_comparison}
\end{figure}

CRNN consists of a ResNet152 and an RNN, where ResNet152 extracts video features and the RNN produces action recognition results. Since ResNet is widely used in video encoders, end-to-end performance comparisons on CRNN demonstrate the generality of our work. We modified CRNN's backbone interfaces to be compatible with SparseTem and tested its acceleration effects on the UCF101 dataset.
Figure \ref{fig:crnn_demo} visualizes CRNN+SparseTem during inference on a video chunk. The model performs dense computation on the 1st frame of the video chunk and performs Diff Computation on subsequent frames. SparseTem can skip the zero pixels, which appear black in the visualization. Accumulation produces the correct feature maps using the sparse feature maps from the last convolution layer. All the feature maps produced by the backbone are then fed into the RNN to obtain the final prediction. 

We conducted end-to-end experiments on three input resolutions: 224, 320, and 420. The batch size was 3, and the temporal length of each video chunk was 28. 
We compared the throughput and accuracy of SparseTem with DeltaCNN and the original model whose backend is cuDNN. The experimental results shown in Figure \ref{fig:crnn_end2end} indicate that our proposed method can optimize the throughput of ResNet152 with more than 4x acceleration while incurring accuracy degradation of less than 1\%. Compared to the SOTA DeltaCNN, SparseTem achieves more than 1.79x acceleration. Figure \ref{fig:location}(a) shows that SparseTem is the most computationally efficient choice when latency is more critical than accuracy.

\begin{table*}[ht]
\centering
\vspace{-20pt}
\begin{tabular}{cccccccc}
\toprule
Model           & Backend    & AP@0.5 (\%) & AP@.5:95 (\%) & \makecell{Latency (ms/frame)} & Speedup & GFLOPS & \makecell{Memory (MB)} \\ \midrule
d4 & cuDNN      & 72.0  & 41.0   & 20.11              & 1       & 52.47   & 796.80       \\
   & DeltaCNN   & 72.0  & 40.8   & 23.79              & 0.85    & 23.74   & 2746.00      \\
   & \textbf{SparseTem}  & 71.8  & 40.4   & \textbf{12.80}              & \textbf{1.57}    & 29.49   & \textbf{492.00}       \\ \midrule
d5 & cuDNN      & 73.0  & 41.1   & 45.31              & 1       & 130.08  & 1243.60      \\
   & DeltaCNN   & 72.7  & 40.7   & 39.20              & 1.16    & 59.81   & 5734.00      \\
   & \textbf{SparseTem}  & 72.6  & 40.4   & \textbf{25.70}              & \textbf{1.76}    & 73.87   & \textbf{899.18}       \\ \midrule
d6 & cuDNN      & 74.1  & 43.7   & 59.16              & 1       & 218.74  & 1593.60      \\
   & DeltaCNN   & 73.3  & 43.1   & 50.16              & 1.18    & 118.68  & 7368.00      \\
   & \textbf{SparseTem}  & 73.1  & 42.7   & \textbf{33.03}              & \textbf{1.79}    & 138.69  & \textbf{1190.33}      \\ \bottomrule
\end{tabular}
\caption{Performance comparison on different levels of Efficientdet and backends.}
\label{tab:performance_comparison}
\end{table*}

\begin{figure}[ht] 
  \centering
  \includegraphics[width=\linewidth]{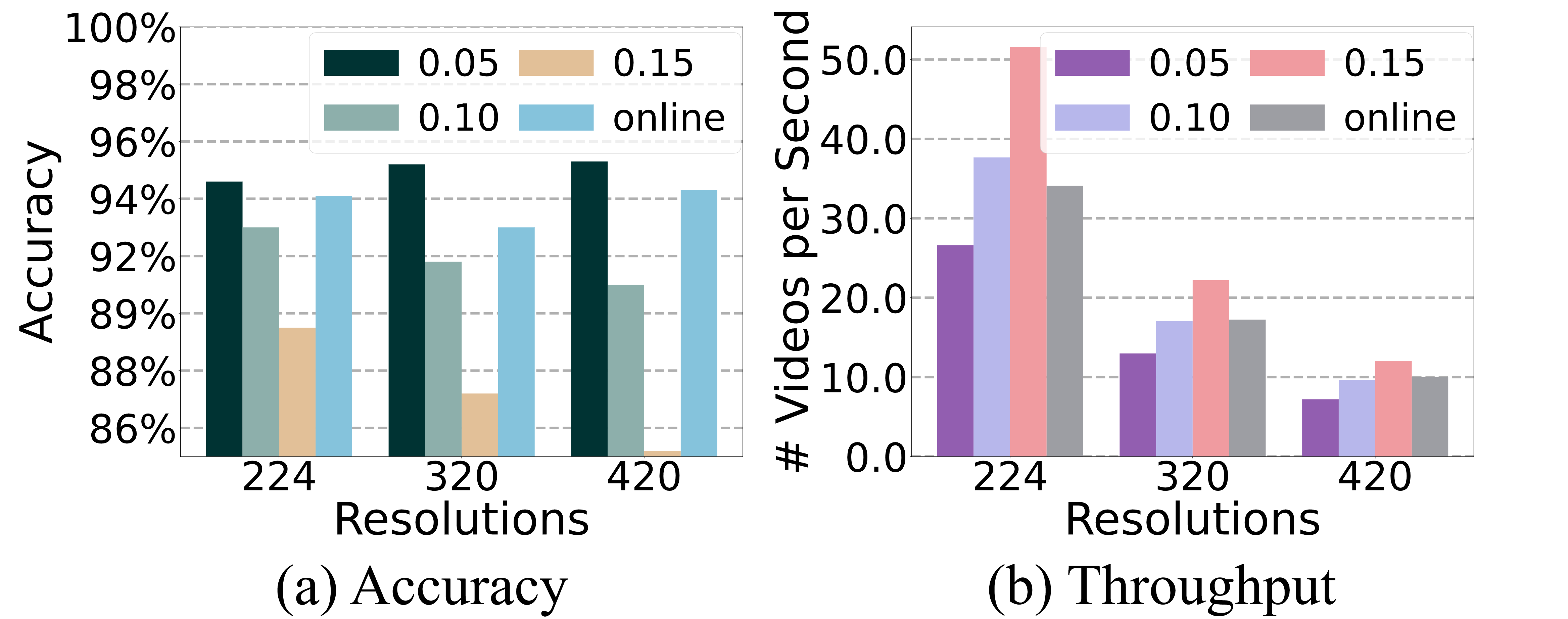}
  \caption{Two folds that illustrate the accuracy and throughput across different truncation thresholds. 
  }
  \label{fig:ablation_online}
\end{figure}

\subsection{Object Detection}

\begin{figure}[ht] 
  \centering
  \includegraphics[width=\linewidth]{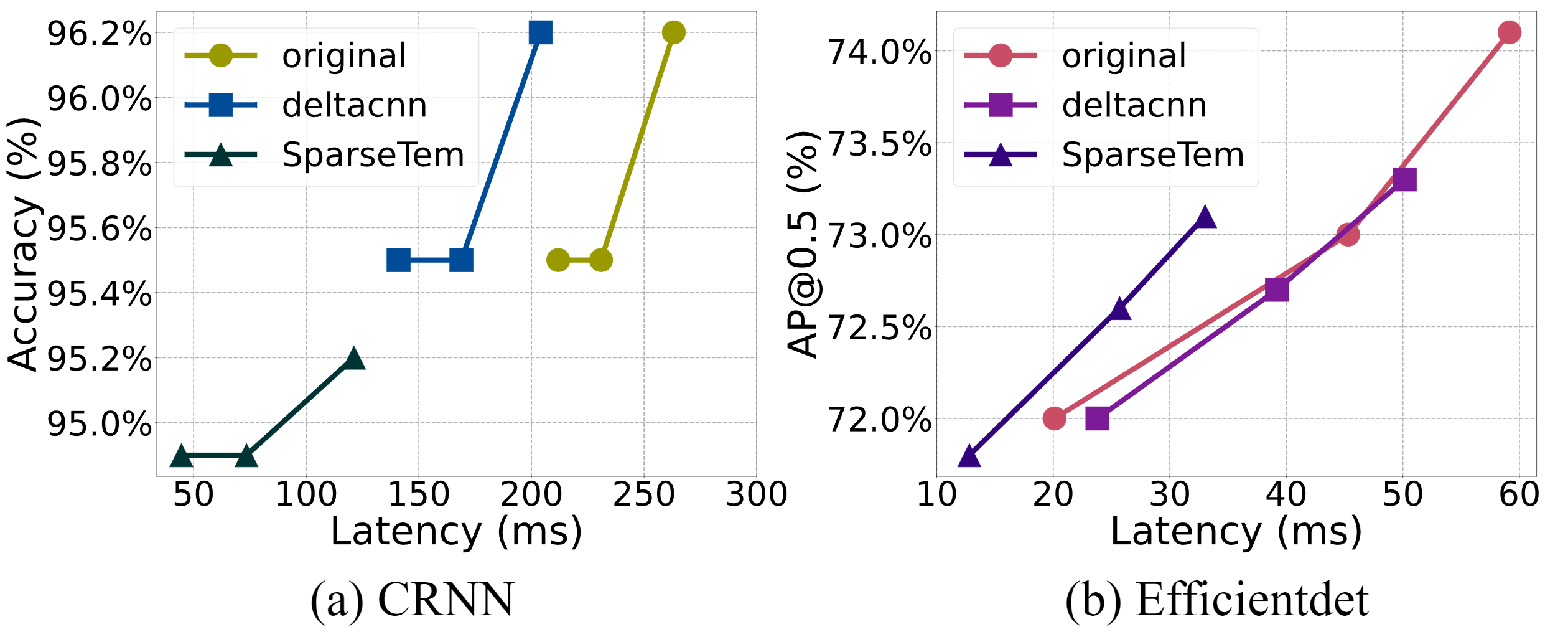}
  \caption{Comparison of SparseTem, the original model, and DeltaCNN on CRNN and EfficientDet across different configurations.
  }
  \label{fig:location}
\end{figure}

EfficientDet \cite{Efficientdet}, which uses EfficientNet \cite{efficientnet} as its backbone, is a top-performing model for object detection, making it ideal for our evaluation. We use Average Precision (AP) as the accuracy metric to compare SparseTem with the original cuDNN-based model and the SOTA DeltaCNN. We compare accuracy, latency, GFLOPS, and per-frame memory cost. The models are a series of EfficientDet, with d0 being the most lightweight. Each model processes one video at a time. SparseTem accelerates inference by 1.57x to 1.79x compared to cuDNN (see Table \ref{tab:performance_comparison}). Although SparseTem saves fewer GFLOPS than DeltaCNN, it still achieves a speedup of 1.51x to 1.85x compared to DeltaCNN. It is worth noting that DeltaCNN consumes significantly more memory than cuDNN and SparseTem due to the burden of caching intermediate feature maps. The accuracy drop of SparseTerm compared to cuDNN is within 1\%, which is acceptable. 
Figure \ref{fig:location}(b) shows that SparseTem significantly reduces latency with a reasonable accuracy drop, making it more computationally efficient than cuDNN and DeltaCNN.


\begin{figure}[ht] 
  \centering
  \includegraphics[width=\linewidth]{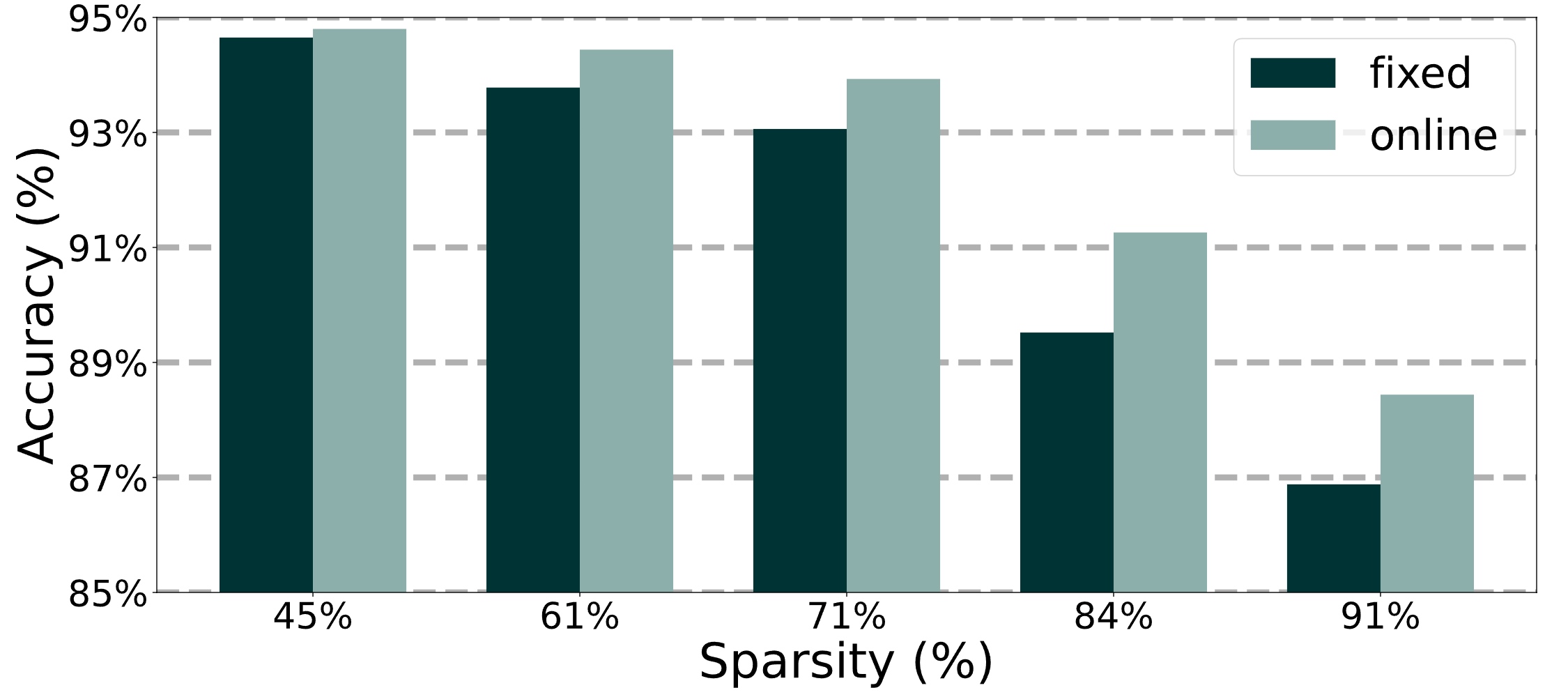}
  \caption{Accuracy comparison between fixed and online threshold truncation across varying sparsity levels.}
  \label{fig:samesparsity}
\end{figure}

\begin{table}[ht]
\centering
\begin{tabular}{cccc}
\toprule
\makecell{Sparsity(\%)} & \makecell{AP@0.5(\%)} & \makecell{AP@.5:95(\%)} & \makecell{Latency(ms)} \\ \midrule
60   & 72.7  & 40.7  & 34.93 \\
70   & 72.7  & 40.7  & 33.60 \\
80   & 72.7  & 40.6  & 31.04 \\
90   & 72.7  & 40.5  & 27.96 \\
95   & 72.6  & 40.4  & 26.83 \\ \bottomrule
\end{tabular}
\caption{Performance comparison at different sparsity levels.}
\label{tab:sparsity_comparison}
\end{table}


\subsection{Ablation Study}
\subsubsection{Memory overhead}
Memory is one of the most important metrics for video encoders. In memory-limited scenarios, it is even more critical than accuracy and latency. We measured the memory cost of the original model, DeltaCNN, and SparseTem on EfficientDet-D3 across various batch sizes. Figure \ref{fig:memory_comparison} shows that DeltaCNN consumes much more memory than the original model and SparseTem. 
With a batch size of 14, DeltaCNN consumes 2.23 times the memory of the original model and 3.1 times that of SparseTem. Despite its computational efficiency, this significant memory overhead limits its practical use. Notably, SparseTem has lower memory overhead than the cuDNN-based original model, proving Sparse Batch's effectiveness and making SparseTem suitable for memory-limited scenarios.

\subsubsection{Effectiveness of truncation threshold adjustment}
Figure \ref{fig:sparsity_diff_fixed_online}(a) shows the imbalance of sparsity across the layers of model. This imbalance can cause obvious accuracy drop because the layer with high sparsity may skip important computation. We propose Online Truncation Threshold Adjustment to dynamically modify the threshold of each layer. Figure \ref{fig:sparsity_diff_fixed_online}(c) proves that the online adjustment can balance the sparsity across layers. To prove that online adjustment can achieve higher accuracy than fixed threshold, we measure the accuracy across different sparsity. Figure \ref{fig:samesparsity} shows that for the same sparsity, online adjustment achieves higher accuracy than a fixed threshold. The accuracy improvement grows with increasing sparsity. At low sparsity, imbalances may lead to higher sparsity in some layers, but it remains within a safe threshold. At high sparsity, however, these imbalances cause a significant drop in accuracy.

\subsubsection{Effect of sparsity}
To find out the effect of sparsity level on accuracy and latency, we use SparseTem to accelerate EfficientDet-D5 under different sparsity levels. Table \ref{tab:sparsity_comparison} shows that with the increasement of sparsity, the latency and accuracy decrease. This is consistent with our intuition. SparseTem can skip more computations when sparsity is higher, leading to greater speedup. However, skipping some important computations may result in an accuracy drop. SparseTem can achieve high accuracy on EfficientDet even when sparsity is 95\%. For CRNN, when sparsity rises from 45\% to 91\%, accuracy drops by 6\%(see Figure \ref{fig:samesparsity}), which is significant. EfficientDet consists of EfficientNet and FPN and we believe the FPN makes EfficientDet robust to high sparsity. CRNN consists of a ResNet152 and a simple RNN, which does not provide CRNN with the same robustness, so high sparsity is not suitable for CRNN.

%% file: sec-conclusion.tex
\section{Conclusions}
 In this work, we propose SparseTem which exploits temporal continuity to boost the efficiency of CNN-based video encoders. we propose SparseBatch to eliminate the need for caching intermediate feature maps in previous work. With Online Truncation Threshold Adjustment, SparseTem can remove the sparsity imbalance across the layers of model to improve the accuracy. 
 Extensive experiments show that our approach can outperform the dense model which uses cuDNN as backend and the SOTA named DeltaCNN within minor accuracy drop. In an era where increasing computational and storage demands are hindering the advancement of video processing, we believe that the computationally efficient and memory-friendly framework presented in this paper can play a pivotal role in this domain.